# Grid Frequency Forecasting in University Campuses using Convolutional LSTM

Aneesh Sathe[1] · Wen-Ren Yang[2]

**Abstract**
The modern power grid is facing increasing complexities, primarily stemming from the integration of renewable energy sources and evolving consumption patterns. This paper introduces an innovative methodology that harnesses Convolutional Neural Networks (CNN) and Long Short-Term Memory (LSTM) networks to establish robust time series forecasting models for grid frequency. These models effectively capture the spatiotemporal intricacies inherent in grid frequency data, significantly enhancing prediction accuracy and bolstering power grid reliability. The research explores the potential and development of individualized Convolutional LSTM (ConvLSTM) models for buildings within a university campus, enabling them to be independently trained and evaluated for each building. Individual ConvLSTM models are trained on power consumption data for each campus building and forecast the grid frequency based on historical trends. The results convincingly demonstrate the superiority of the proposed models over traditional forecasting techniques, as evidenced by performance metrics such as Mean Square Error (MSE), Mean Absolute Error (MAE), and Mean Absolute Percentage Error (MAPE). Additionally, an Ensemble Model is formulated to aggregate insights from the building-specific models, delivering comprehensive forecasts for the entire campus. This approach ensures the privacy and security of power consumption data specific to each building.

## 1 Introduction

The global energy landscape is undergoing a profound transformation due to increasing demand for sustainable and reliable power sources. As our societies become more reliant on electricity, the significance of a stable and well-regulated power grid cannot be emphasized enough. At the heart of this electrified world lies the intricate and dynamic power grid, serving as the backbone of modern civilization. Within this extensive network, one critical parameter emerges as the pivotal measure of grid health and stability: Grid Frequency.

Grid Frequency, often maintained at a standard value like 50Hz or 60Hz, serves as an indicator of the equilibrium between electricity generation and consumption within the power grid.

[1] School of Computer Science & Engineering, Vellore Institute of Technology, Chennai, India.

[2] Electric Engineering Dept., National Changhua University of Education, Changhua, Taiwan.

It essentially measures how quickly alternating current (AC) electricity oscillates back and forth. This frequency is tightly regulated to ensure that it remains within a very narrow range around the standard value. Grid Frequency is vital for the proper functioning of the power grid. It ensures that the balance between electrical supply and demand is maintained, and any deviation from this standard frequency can disrupt daily operations and inconvenience consumers.

As the power grid becomes increasingly complex with the integration of renewable energy sources, distributed generation, and varying consumption patterns, the need for precise and efficient grid frequency forecasting becomes a necessity. Traditional forecasting methods have proven inadequate in capturing the sophisticated dynamics of modern power grids, which often exhibit nonlinear and ever-changing behaviour.

This paper explores the fusion of Convolutional Neural Networks (CNN) and Long Short-Term Memory (LSTM) neural networks to create a robust and accurate time series forecasting model for grid frequency. By harnessing the spatial and temporal features inherent in grid frequency data, this innovative approach aims to boost our capability to predict and mitigate frequency deviations, ultimately enhancing the reliability and proper functioning of the power grid.

The usefulness of such a forecasting system spans a broad array of applications, including grid management and control, optimization of energy markets, and the integration of renewable energy sources. These are all critical components in the shift towards a sustainable and efficient energy ecosystem.

## 2 Related Work

The authors [1] investigate the intricate dynamics of electrical power systems' frequency, a critical factor in ensuring grid stability. It emphasizes the need for accurate frequency predictions due to the hybrid nature of power grid frequency—comprising both stochastic and deterministic influences.

The study employs machine learning, specifically k-weighted-nearest-neighbour (WNN) methods, for forecasting power grid frequency. What sets this research apart is its enhancement of the methodology, achieving superior prediction accuracy and shorter forecasting horizons. This improvement has significant implications for grid stability and cost savings. Importantly, the study extends the methodology's applicability, making it adaptable to various power systems. This opens doors for widespread use, enhancing the stability and reliability of diverse power grids.

The Authors of paper [2] addresses the challenges posed by the growing integration of inverter-based renewable energy sources, which profoundly affect power system inertia and frequency stability. The research explores five machine learning methods for predicting frequency nadir—a crucial parameter for pre-empting frequency deviations. Notably, the study focuses on gradient boosting and XGBoost as the top-performing methods.

It showcases their effectiveness in predicting nadir frequencies, particularly in scenarios characterized by high levels of renewable energy penetration. The technical prowess of these machine learning methods in handling complex power system data is highlighted, making them invaluable tools for power grid operators and planners.

The paper [3] dives into the feasibility of predicting Electrical Network Frequency (ENF) values—a vital aspect of power grid management. The authors employ advanced techniques such as auto-correlation and correlation coefficient analyses to forecast ENF values accurately. Notably, the research proposes two distinct forecasting approaches: kernel regression with correlation coefficient and autoregressive moving average models.

Through meticulous experimentation on ENF data from three U.S. power grids, the study reveals the potential of precise ENF predictions to bolster power grid reliability. This research offers a technical roadmap for power grid operators seeking to enhance grid stability through advanced data analysis and prediction methods.

Another research [4] introduces a visionary approach to Frequency Stability Prediction (FSP) using vision transformers (ViTs). ViTs process time-series data from power system operations in real-time, aiding operators in swiftly mitigating losses following power disturbances.

The technical innovation lies in ViT's capacity for high-dimensional data processing and global feature extraction from power system operations. Moreover, the research incorporates copula entropy-based feature selection to streamline data processing and eliminate redundancy, making it computationally efficient.

The introduction of a Frequency Security Index (FSI) as a prediction indicator enhances the technical sophistication of the approach. Simulations demonstrate the ViT-based FSP's superiority, positioning it as a state-of-the-art solution for safeguarding power grid stability in the face of evolving energy landscapes.

The paper [5] delves into the critical realm of precise solar power forecasting, especially in the context of grid-integrated distributed energy resources (DERs). It conducts a comprehensive review of various artificial neural network (ANN) architectures typically employed in power forecasting, with a focal point on the adeptness of Long Short-Term Memory (LSTM) networks.

The study [5] introduces its novel approach, employing a streamlined LSTM model encapsulated within a machine learning framework tailored for one day-ahead solar power prediction. Experimental findings, leveraging data from two photovoltaic (PV) sites, unequivocally underscore the supremacy of LSTM over Multilayer Perceptron (MLP) in forecasting tasks.

Furthermore, the paper scrutinizes the influence of weather conditions on forecasting precision and conducts an in-depth exploration of hyperparameter tuning. Future research horizons are envisioned to encompass multi-variate time series forecasting, necessitating the assimilation of meteorological data.

## 3 Methodology

Fig 1. Illustrates the layout of the Three Engineering Buildings in the Bao-Shan campus of NCUE, Taiwan. The aim is to develop robust prediction models, specific to each engineering buildings and an overall prediction model for the entire campus using an ensemble of the building-specific models.

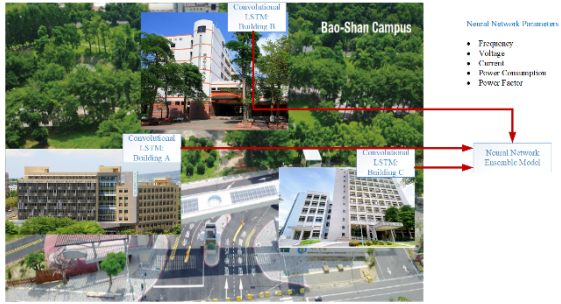

**Fig. 1** Layout of the Bao-Shan Campus

Fig 2. provides a visual representation of the methodology, outlining the sequence of steps involved in predicting grid frequency for the entire campus. Each box in the diagram corresponds to a specific task or stage in the process, and arrows indicate the flow of data and results between these tasks.

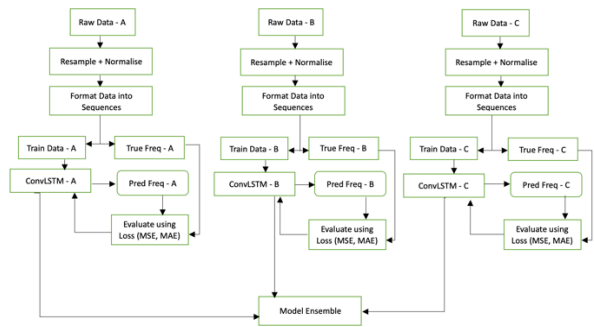

**Fig. 2** Prediction Model Flowchart

Our approach involves developing an ensemble model without direct access to the training data for each of the three engineering buildings on campus. The process begins with the collection of raw data specific to each building, which is then stored as separate CSV files. To ensure uniformity, the raw data is resampled to a one-minute frequency, overcoming the challenge of non-uniform data collection. Subsequently, data normalization is applied, and feature engineering is performed to select the most pertinent features for model training. The formatted data is then organized into batches of sequences, consisting of input features (X) and the target grid frequency values (y) for training the ConvLSTM models for each individual building.

Building-specific ConvLSTM models are developed and trained independently, with model evaluations based on performance metrics such as Mean Square Error (MSE). Finally, an ensemble model for the entire campus is created by combining the parameters of the individual building-specific models, enabling accurate predictions of grid frequency for the entire campus while maintaining data confidentiality.

## 4 Theory

### 4.1 Long-Short Term Memory (LSTM)

Recurrent Neural Networks (RNNs) are a class of artificial neural networks specifically designed for processing sequences of data. Unlike traditional feedforward neural networks, which take fixed-sized inputs and produce fixed-sized outputs, RNNs are capable of handling input sequences of varying lengths. At the core of an RNN is a hidden state that evolves as the network processes each element of the input sequence. The fundamental idea is to maintain a form of "memory" within the network that retains information about the previous elements in the sequence, allowing the network to capture temporal dependencies.

While RNNs are powerful for modelling sequences, they suffer from a significant limitation known as the "vanishing gradient" problem. As gradients are propagated backward through time during training, they can become extremely small when dealing with long sequences. This causes the network to have difficulty learning and capturing dependencies over extended time intervals.

To address the vanishing gradient problem and enhance the modelling of long-range dependencies, (Long Short-Term Memory) LSTMs were introduced. LSTMs are a specialized type of RNN architecture that incorporates memory cells and gating mechanisms. These features enable LSTMs to selectively remember, update, or forget information, making them highly effective at modelling sequences with extended temporal context. This makes them well-suited for tasks involving sequential data, such as time series forecasting, natural language processing, and speech recognition.

LSTMs [10] employ three essential gates: the Forget Gate ($f_t$), the Input Gate ($i_t$), and the Output Gate ($O_t$). These gates play a central role in regulating information flow and memory management within the LSTM. Alongside the gates, a Cell State ($C_t$), Candidate Cell State ($\bar{C}_t$), and a Hidden State ($h_t$) is also maintained.

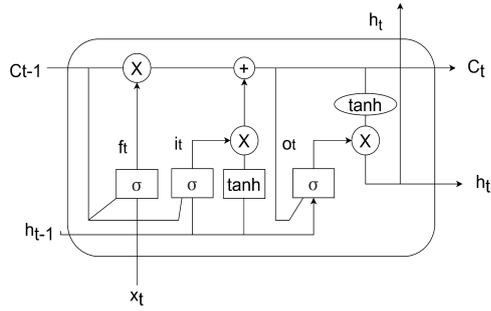

**Fig. 3** Single LSTM Cell

The mathematical underpinnings of an LSTM at each timestep 't' involve a series of operations :

$$f_t = \sigma(W_f \cdot [h_{t-1} \cdot x_t] + b_f)$$
$$i_t = \sigma(W_i \cdot [h_{t-1} \cdot x_t] + b_i)$$
$$\bar{C}_t = \tanh(W_c \cdot [h_{t-1} \cdot x_t] + b_C)$$
$$C_t = (F_t * C_{t-1} + i_t * \bar{C}_t)$$
$$O_t = \sigma(W_o \cdot [h_{t-1} \cdot x_t] + b_o)$$
$$h_t = o_t * \tanh(C_t)$$

(1)

**Table 1** Variables in an LSTM Cell

| Variable | Definition |
| --- | --- |
| $x_t$ | Input |
| $[h_{t-1} \cdot x_t]$ | Concatenation of the previous hidden state and the current input |
| $\sigma$ | Sigmoid Activation |
| $W_x$ | Weight matrix specific to gate 'x' |
| $b_x$ | Bias term specific to gate 'x' |

### 4.2 One Dimensional Convolution (Conv1D)

The Conv1D layer [9] is a mathematical operation designed for uncovering patterns within one-dimensional sequences, making it suitable for processing time series data, audio signals, or textual information. It operates by applying a set of adaptable filters, referred to as kernels, to the input sequence to identify pertinent features.

Within the Conv1D layer, a collection of filters is employed, each equipped with its own learnable weights. These filters traverse the input sequence, performing a weighted dot product with the local data at each position. The outcome is a feature map that encodes the presence of specific patterns and features within the sequence. Stride determines the incremental step size by which the filter traverses the input sequence. A larger stride leads to a reduction in the spatial resolution of the output feature map, potentially accelerating computation. Conversely, a smaller stride allows for the capture of finer, more detailed information.

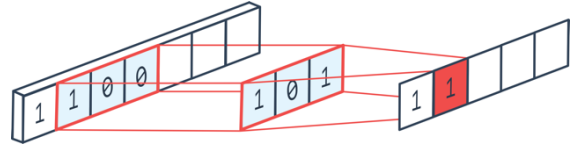

**Fig. 4** Conv1D Operation

In order to control the spatial dimensions of the resulting feature map, Padding is utilized. The introduction of zero-padding involves appending zeros to the edges of the input sequence before the convolution operation. This practice preserves the spatial size of the feature map and helps alleviate issues associated with information loss at the sequence boundaries.

## 5 Experiment

### 5.1 Dataset

The Raw Data was collected for each of the three engineering buildings over a span of one month from "2022-10-01" to "2022-11-01" and stored as a CSV File. The Raw Data has 29 Features, from which 8 features were selected for training the model. The Selected Features are as follows :

**Table 2** Features for the Prediction Model

| Notation | Feature |
| --- | --- |
| UpdateTime | Timestamp of Recorded Data |
| $I_a$ | Three Phase Current - A |
| $I_b$ | Three Phase Current - B |
| $I_c$ | Three Phase Current - C |
| $V_a$ | Three Phase Voltage - A |
| $V_b$ | Three Phase Voltage - B |
| $V_c$ | Three Phase Voltage - C |
| PF | Power Factor |
| Freq | Frequency (Hz) |

The Root Mean Square (RMS) Value of the Three Phase Currents and Three Phase Voltages is considered, and recorded intermittently in the Raw Dataset.

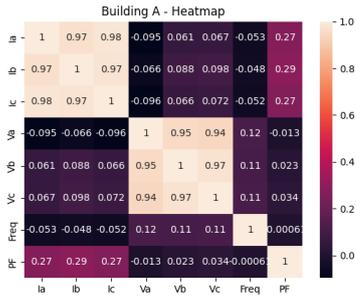

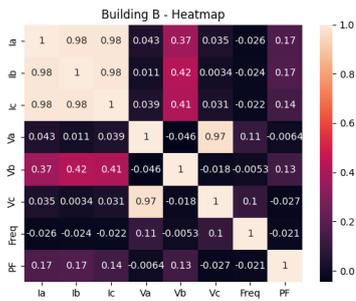

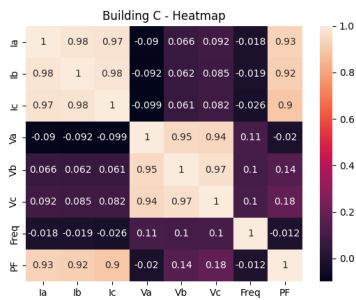

**Fig. 5** Feature Correlation Heatmaps of Data for each individual building

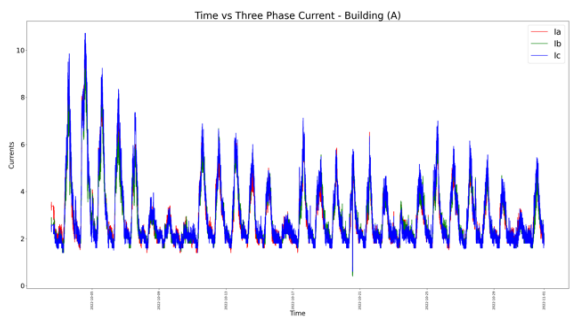

**Fig. 6** Three Phase Current – Building A

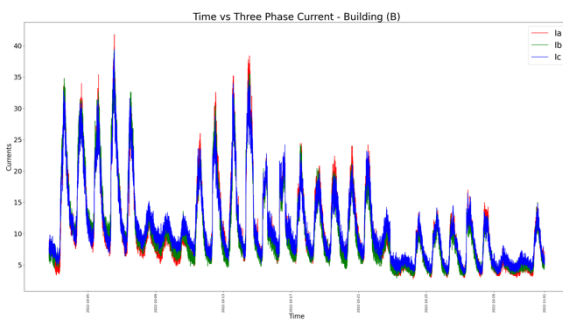

**Fig. 7** Three Phase Current – Building B

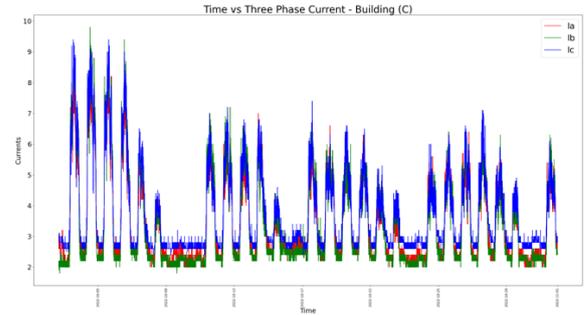

**Fig. 8** Three Phase Current – Building C

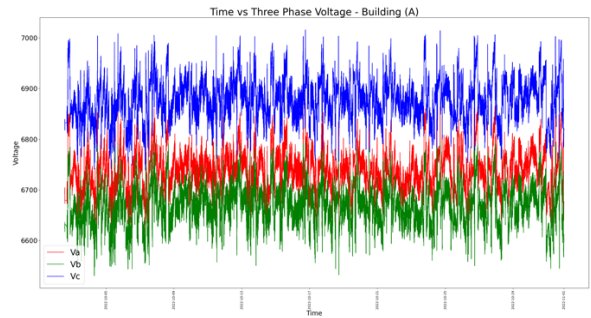

**Fig. 9** Three Phase Voltage – Building A

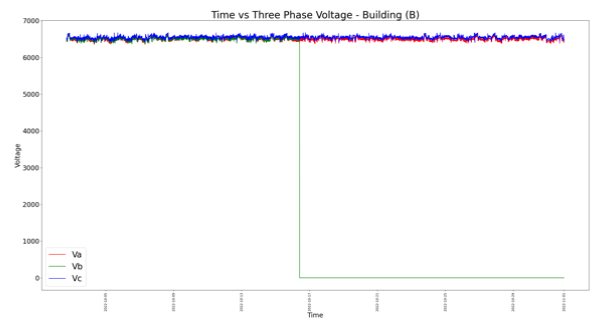

**Fig. 10** Three Phase Voltage – Building B

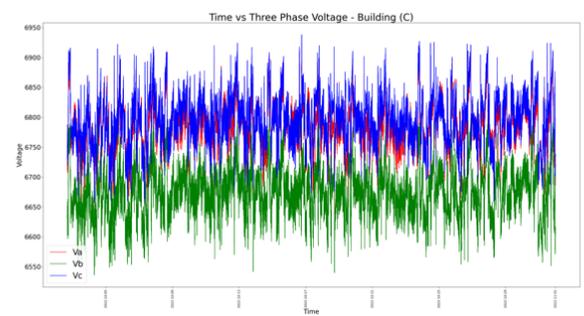

**Fig. 11** Three Phase Voltage – Building C

Some unexpected characteristics can be observed when looking at the three phase voltage data. Some magnitude shifts can be seen in Fig. 7 due to Imbalanced loads. The loads connected to the system may not be evenly distributed throughout the phases, some phases ($V_a$ and $V_b$) may be required to bear a greater load than the others ($V_c$). This results in voltage decrease in the overloaded phase, resulting in a decreased voltage value. These kinds of unique characteristics as mentioned in the works [6] and [8] is what makes this dataset valuable as simulated or hypothetical data might not consider these real-life situations therefore, compromising the accuracy of developed prediction model in the real-world application.

## 5.2 Data Preprocessing

The raw data undergoes a crucial preprocessing phase to address various challenges such as non-uniform sampling frequency, missing values, and potential measurement errors. This preprocessing step is essential to establish data uniformity and prepare it for effective use in our prediction models.

Initially, the raw data is resampled to achieve a consistent sampling interval of one minute. This resampling process ensures that each of the individual buildings' datasets shares a uniform frequency, eliminating disparities caused by irregular sampling intervals.

Subsequently, a normalization technique is applied to the data. This normalization procedure, which can involve methods like Min-Max Scaling, transforms the data into a standardized range of values spanning from 0.0 to 1.0. This standardized format enhances the efficiency of training for each ConvLSTM model, ensuring that the models can effectively learn and generalize from the preprocessed data.

## 5.3 Data Preparation

In order to effectively train Long Short-Term Memory (LSTM) models, the acquired data undergoes a crucial formatting process to enable accurate feature mapping. The post-processed data is reshaped into sequences suitable for LSTM-based supervised learning. Each sequence is structured as a 3D tensor characterized by dimensions comprising batch size, sequence length, and feature dimension.

This reshaping procedure is pivotal for casting the problem as a supervised learning task for LSTM, wherein the objective is to predict the grid frequency at the current timestep [t0]. This prediction relies on utilizing data features (including $I_a$, $I_b$, $I_c$, $V_a$, $V_b$, $V_c$, PF, and Frequency) from previous timesteps [$t_{n-1}$, $t_{n-2}$, … $t_{-2}$, $t_{-1}$].

**Table 3** Input Sequence Length for each model

| Model | Input Sequence | Output Sequence |
|---|---|---|
| ConvLSTM-A | 7 | 1 |
| ConvLSTM-B | 5 | 1 |
| ConvLSTM-C | 3 | 1 |

Table 3. outlines the specific input sequence lengths for each ConvLSTM model employed in the study: ConvLSTM-A: Receives an input sequence of length 7, enabling it to predict the grid frequency at the subsequent timestep.

ConvLSTM-B: Processes an input sequence of length 5 to make predictions about the grid frequency at the next timestep.

ConvLSTM-C: Accepts an input sequence of length 3 and produces predictions for the grid frequency at the next timestep.

## 5.4 ConvLSTM Model Specifications

The ConvLSTM (Convolutional Long Short-Term Memory) is an advanced neural network architecture that combines the strengths of Convolutional Neural Networks (CNNs) and Long Short-Term Memory (LSTM) networks. This fusion enables ConvLSTMs to capture both spatial and temporal dependencies in sequential data, making them particularly effective for time-series forecasting.

The specifications for each prediction model used for forecasting Grid Frequency in each of the buildings are organised in the table below:

**Table 4** Model Specifications

| Model Architecture | ConvLSTM A | ConvLSTM B | ConvLSTM C |
|---|---|---|---|
| Input Layer | 3D tensor with dimensions (batch size, sequence length, feature dimension) | 3D tensor with dimensions (batch size, sequence length, feature dimension) | 3D tensor with dimensions (batch size, sequence length, feature dimension) |
| Conv1D Layer | In-Channels: 7 Out-Channels: 64 Kernel: 3 Padding: 1 Stride : 1 | In-Channels: 5 Out-Channels: 64 Kernel: 3 Padding: 1 Stride : 1 | In-Channels: 3 Out-Channels: 32 Kernel: 3 Padding: 1 Stride : 1 |
| ReLU Activation | | | |
| LSTM Layer | Input Size: 8 Hidden Size: 32 No. Classes: 1 Dropout : 0.1 | Input Size: 8 Hidden Size: 32 No. Classes: 1 Dropout : 0.1 | Input Size: 8 Hidden Size: 32 No. Classes: 1 Dropout : 0.1 |
| ReLU Activation | | | |
| Fully Connected Layer 1 | Input Size : Hidden Size Output Size : 10 | Input Size : Hidden Size Output Size : 10 | Input Size : Hidden Size Output Size : 10 |
| ReLU Activation | | | |
| Fully Connected Layer 2 | Input Size : 10 Output Size : 1 | Input Size : 10 Output Size : 1 | Input Size : 10 Output Size : 1 |
| Output Layer | Frequency Prediction | Frequency Prediction | Frequency Prediction |
| **Hyperparameters** | | | |
| Loss Function | Mean Square Error, Mean Absolute Error | Mean Square Error, Mean Absolute Error | Mean Square Error, Mean Absolute Error |

| Optimiser | ADAM | ADAM | ADAM |
| Learning Rate | 0.0001 | 0.0001 | 0.0001 |
| Training Epochs | 1500 | 1500 | 2000 |

Each model has been carefully tailored to capture the temporal dependencies, context, and patterns within the data. Key architectural elements, such as Conv1D layers, LSTM layers, fully connected layers, and output layers, have been detailed to elucidate the flow of information and feature extraction within the models.

Hyperparameters, including sequence length, hidden units, learning rates, and training epochs, have been judiciously chosen or experimentally determined to strike a balance between computational efficiency and model convergence. We have also incorporated regularization techniques, such as dropout in the LSTM layers, to prevent overfitting during training.

## 6 Results

The trained ConvLSTM Models are evaluated using Mean Square Error (MSE) Loss and Mean Absolute Error (MAE) Loss. The Train-Test Loss Curve for each of the prediction models can be graphically with 'Number of Epochs' as x-axis and 'Loss' as y-axis.

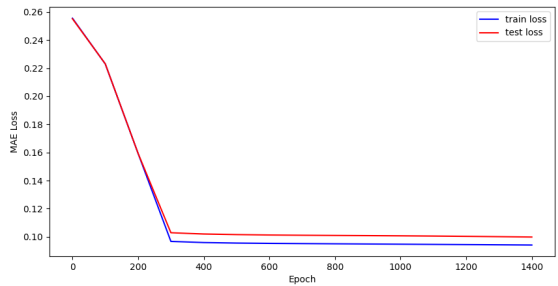

**Fig. 12** MAE Loss ConvLSTM – A

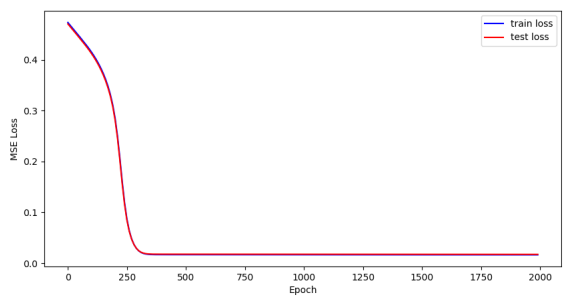

**Fig .13** MSE Loss ConvLSTM – A

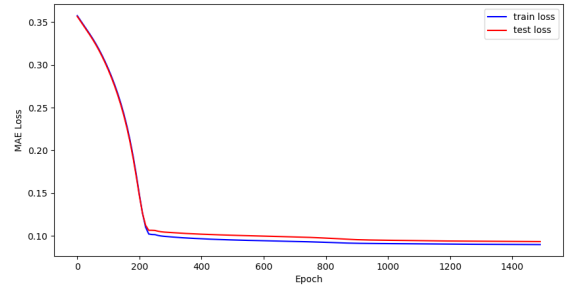

**Fig. 14** MAE Loss ConvLSTM – B

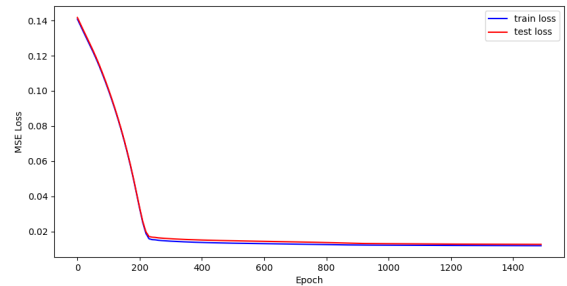

**Fig. 15** MSE Loss ConvLSTM – B

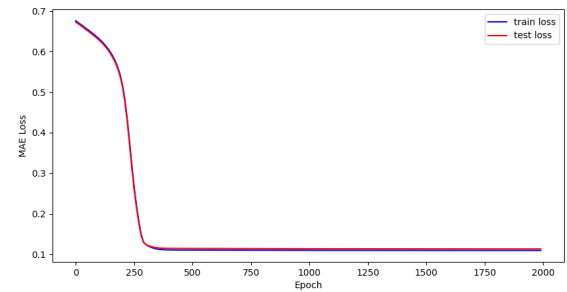

**Fig. 16** MAE Loss ConvLSTM – C

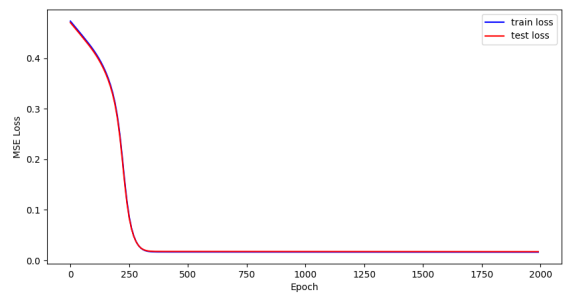

**Fig. 17** MSE Loss ConvLSTM – C

The frequency predictions (in Hertz) from the models were compared with the true frequency values (in Hertz) from the data from respective buildings as a function of Time. The Actual vs Predicted Frequency for each of the Three building-specific prediction models can be graphically visualised as:

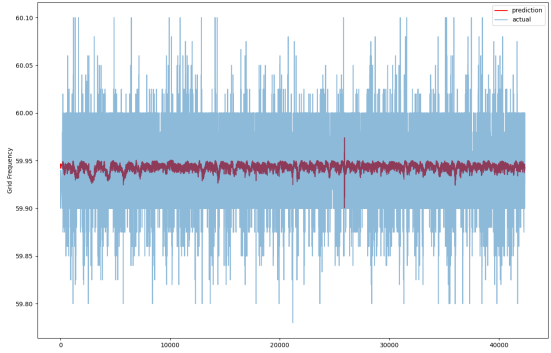

**Fig. 18** Actual vs Prediction – ConvLSTM A

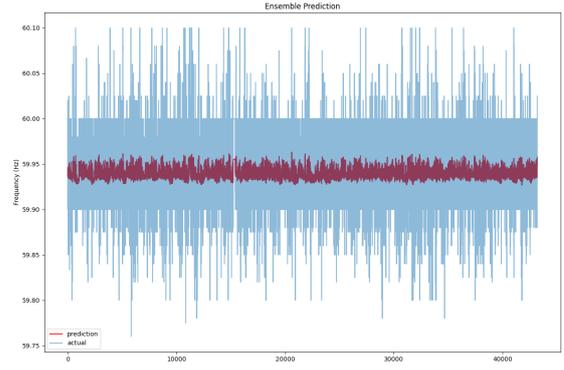

**Fig. 21** Actual vs Prediction – Ensemble Model

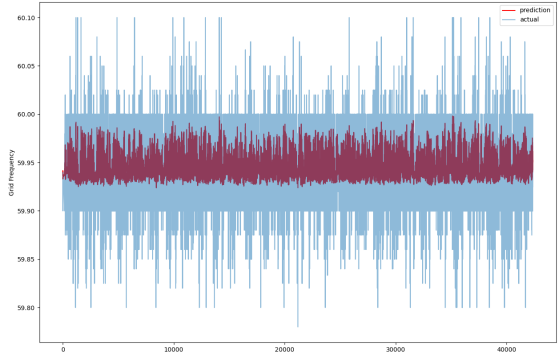

**Fig. 19** Actual vs Prediction – ConvLSTM B

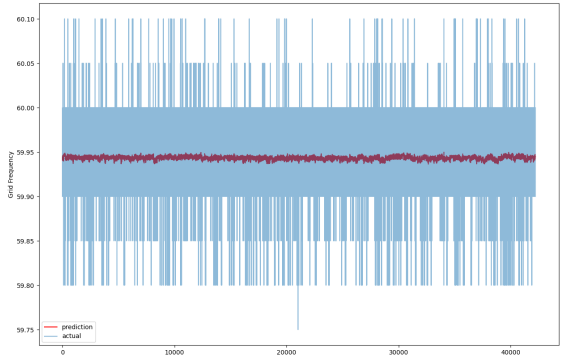

**Fig. 20** Actual vs Prediction – ConvLSTM C

Mean Square Error (MSE) is a metric used to assess the accuracy of predictive models. It calculates the average of the squared differences between predicted and actual values.

$$MSE = \sum_{i=1}^{N} \frac{(Predicted_i - Actual_i)^2}{N} \qquad (7)$$

Mean Absolute Error (MAE) is a metric for evaluating model accuracy. It computes the average of the absolute differences between predicted and actual values.

$$MAE = \sum_{i=1}^{N} \frac{|Predicted_i - Actual_i|}{N} \qquad (8)$$

Mean Absolute Percentage Error (MAPE) is a metric for assessing prediction accuracy in terms of percentage errors. It calculates the average percentage difference between predicted and actual values.

$$MAPE = \frac{1}{N} \sum_{i=1}^{N} \left| \frac{(Predicted_i - Actual_i)}{Actual_i} \right| \qquad (9)$$

The MSE, MAE and MAPE Loss associated with each of these Models on New Data (Not Previously seen by the prediction models) can be tabulated as :

Additionally, an Ensemble Model is trained from building-specific prediction models to make generalised predictions for the overall campus. New Data (Not Previously seen by the prediction models) is pre-processed in the exact same way as discussed in the preceding sections and passed through each of the building-specific models to obtain predictions. A weighted-average (Table 5) of these predictions is taken and used as an Ensemble Prediction for the entire campus.

**Table 5** Ensemble Weights

| Model | Weight |
| --- | --- |
| ConvLSTM A | 0.3 |
| ConvLSTM B | 0.4 |
| ConvLSTM C | 0.3 |

**Table 6** Model Evaluation

| Loss Metric | ConvLSTM A | ConvLSTM B | ConvLSTM C | Model Ensemble |
| --- | --- | --- | --- | --- |
| MSE | 0.001605 | 0.001414 | 0.001535 | 0.001400 |
| MAE | 0.032767 | 0.030919 | 0.032509 | 0.031074 |
| MAPE | 0.054652 | 0.051574 | 0.054233 | 0.051834 |

## 7 Conclusion

The study highlights that LSTM Based Prediction Models with a Convolutional Layer can be utilized for specific as well as generalised predictions using

Ensembles for Time Series Forecasting like Grid Frequency, which provide a much accurate prediction on average, when compared with using any one of the specific models for overall prediction tasks.

**Acknowledgements** This research is supported by Artificial Neural Network Applications, TEEP (Taiwan Experience Education Program), Taiwan Ministry of Education, Project No. 1110069853.

**Data availability** The dataset used in this study was derived from Power Consumption data analysed in Bao-Shan Campus of NCUE, Taiwan. The data can be accessed here : https://github.com/aneesh-sathe/ncue-power-data

### Declarations

**Conflict of interest** The authors declare no conflict of interest.